\pdfoutput=1

\documentclass[11pt]{article}

\usepackage[final]{acl}

\usepackage{times}
\usepackage{latexsym}
\usepackage{amsmath}
\usepackage{amssymb}
\usepackage{CJKutf8}
\usepackage{arydshln}
\usepackage{ulem}
\usepackage{soul}
\usepackage{subcaption}
\usepackage{cleveref}
\usepackage{soul}
\usepackage{xcolor}
\usepackage{transparent}

\usepackage[T1]{fontenc}

\usepackage[utf8]{inputenc}

\usepackage{microtype}

\usepackage{inconsolata}

\usepackage{graphicx}

\title{Spelling-out is not Straightforward:\\ LLMs' Capability of Tokenization from Token to Characters}

\author{
    Tatsuya Hiraoka \quad\quad Kentaro Inui \\
    Mohamed bin Zayed University of Artificial Intelligence (MBZUAI)   \\
    RIKEN\\
    \texttt{\{tatsuya.hiraoka, kentaro.inui\}@mbzuai.ac.ae}
}

\begin{document}
\maketitle
\begin{abstract}
Large language models (LLMs) can spell out tokens character by character with high accuracy, yet they struggle with more complex character-level tasks, such as identifying compositional subcomponents within tokens.
In this work, we investigate how LLMs internally represent and utilize character-level information during the spelling-out process.
Our analysis reveals that, although spelling out is a simple task for humans, it is not handled in a straightforward manner by LLMs.
Specifically, we show that the embedding layer does not fully encode character-level information, particularly beyond the first character.
As a result, LLMs rely on intermediate and higher Transformer layers to reconstruct character-level knowledge, where we observe a distinct “breakthrough” in their spelling behavior.
We validate this mechanism through three complementary analyses: probing classifiers, identification of knowledge neurons, and inspection of attention weights.
\end{abstract}

\section{Introduction}
\label{sec:introduction}
While large language models (LLMs) have grown remarkably in recent years, several studies report that they still struggle with fine-grained character-level manipulations, such as inserting, deleting, or extracting individual characters within tokens~\cite{edman-etal-2024-cute, wang2024stringllm, chai-etal-2024-tokenization, shin2024large}. 
Although most LLMs operate over subword tokens, true mastery of subtoken information is essential for a range of applications, such as morphological inflection~\cite{marco-fraser-2024-subword}, letter counting~\cite{fu2024large}, typoglycemia~\cite{wang2025word}, and handling typos~\cite{tsuji2025investigating}. 
To improve their reliability in such scenarios, we must understand how LLMs internally represent and process characters.

A paradox emerges from prior work: LLMs can accurately spell out entire tokens as sequences of characters~\cite{edman-etal-2024-cute, xiong2025enhancing}, while they often fail at simple tasks such as identifying a single character at a fixed position~\cite{hiraoka2024knowledge, chai-etal-2024-tokenization}.
For instance of a token ``\texttt{language},'' models readily generate ``\texttt{l a n g u a g e}'' on demand but cannot reliably extract ``\texttt{u}'' at position five. 
This discrepancy suggests that, despite lacking explicit access to compositional character knowledge, LLMs have some mechanism for character-by-character spelling.

\begin{figure}[t]
\centering
\includegraphics[width=7.0cm]{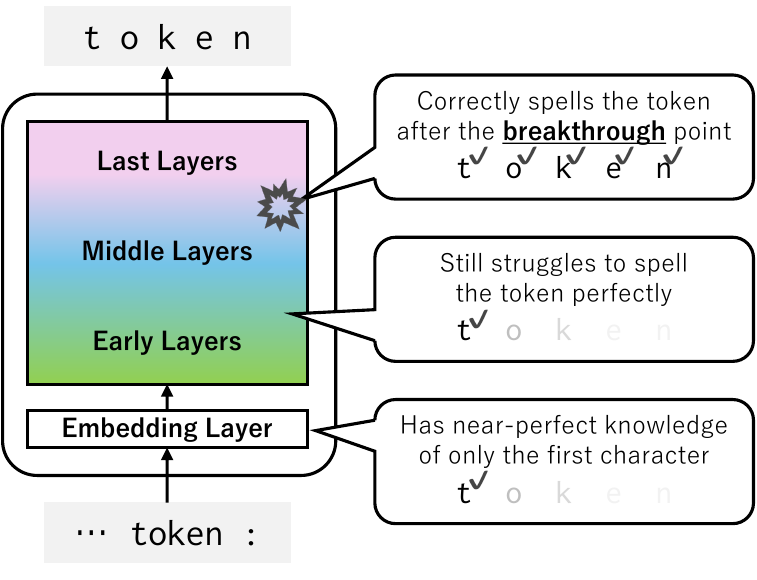}
\caption{
        Summary of our findings. LLMs spell out tokens by relying directly on embedding-level information for the first character, but gradually shift to using distributed, higher-layer representations for later characters. We found a ``breakthrough'' layer where character knowledge becomes detectable.
}
\label{fgr:fewshotResult}
\end{figure}

In this paper, we investigate how and where LLMs capture and deploy character‐level knowledge during spelling-out. 
We begin by constructing a token-characters dataset from the vocabularies of four representative LLMs (\S \ref{sec:settings}) and confirming their ability to spell out tokens with few‐shot prompts (\S \ref{sec:preliminary}). 
Probing the embedding layer reveals that it does not encode subtoken characters directly (\S \ref{sec:embedProbing}), but downstream Transformer layers gradually recover this information. 
We identify a clear ``breakthrough'' layer at which character identities become reliably detectable by our probing classifier (\S \ref{sec:layerProbe}), and observe that the first character is handled differently from subsequent ones. 
Finally, through neuron‐level analyses (\S \ref{sec:neuron}) and attention‐weight inspections (\S \ref{sec:attention}), we trace how character knowledge is stored and routed, demonstrating that its locus aligns precisely with the breakthrough point.

Taken together, our study sheds new light on the internal character-level machinery of LLMs. 
Our main contributions are as follows:
\vspace{-0.6\baselineskip}
\begin{itemize}
    \setlength{\parskip}{0.1cm} 
    \setlength{\itemsep}{0cm} 
    \item We demonstrate that token embeddings do not fully encode character-level information, and that subsequent Transformer layers play a major role in reconstructing this information during the spelling-out process.
    \item We pinpoint the exact layer where compositional character information ``breaks through,'' validated via probing classifiers.
    \item We investigate character knowledge to specific neurons and attention patterns, showing alignment with the breakthrough layer.
    \item We provide a diagnostic framework (dataset, probes, and analyses) for future research into subtoken and character-level modeling.
\end{itemize}



\section{Related Work}
\label{sec:relatedwork}
Our research is in line with work focusing on the LLMs' capability of character-level manipulation.
\citet{edman-etal-2024-cute} introduced a dataset to evaluate this capability from some viewpoints of character-level manipulation, including spelling-out of words or tokens.
Similarly, \citet{wang2024stringllm} provided a dataset of complicated character-level manipulation tasks.
Both works conclude that LLMs have limited capability for complicated character processing.
Furthermore, these works reported that LLMs can spell out words in the original order, as reported in \citet{xiong2025enhancing}, while they do not have sufficient knowledge of their single characters inside words~\cite{shin2024large,hiraoka2024knowledge,chai-etal-2024-tokenization} unless fine-tuning for the models to learn the token internal information directly~\cite{xu2024enhancing}.

This line of literature motivates us to investigate the LLMs' internal workings of spelling-out behavior, despite the lack of character-level knowledge.
Recently, we can see a trend of understanding LLMs' capability of recognizing character-level information, such as the ability of counting characters~\cite{fu2024large}.
Moreover, beyond the word-to-character spelling out, \citet{wu-etal-2025-impact} investigates their ability to recognize radicals inside Chinese characters.

Our work is also related to the findings of LLMs' ``detokenization'' ability in their later layers~\cite{kaplan2025from,kamoda-etal-2025-weight}, which is an ability to internally merge tokens into words or phrases.
In contrast, we focus on the inverse problem: how LLMs internally ``tokenize'' tokens into characters.

\section{Experimental Setup}
\label{sec:settings}

\subsection{Target Models}
We investigate four medium-sized LLMs ($\approx 7$B parameters): LLaMA3-8B~\cite{dubey2024llama}, Gemma-7B~\cite{team2024gemma}, Qwen2.5-7B~\cite{yang2024qwen2}, and Amber~\cite{liullm360}.
We selected them because our preliminary trials showed that smaller models (e.g., 3B) fail to spell out tokens with sufficient accuracy. 
All models have the Transformer-based architecture~\cite{vaswani2017attention}.
Table \ref{tbl:dataset} shows each model’s vocabulary size, which ranges from 32K to 256K tokens.

\subsection{Evaluation Dataset}
\label{sec:dataset}
\begin{table}[]
\small
\centering
\begin{tabular}{lrrr}
\hline 
           & \textbf{\# Vocab} & \textbf{\# Dataset} & \textbf{\%} \\\hline
LLaMA3-8B  & 128,256 & 19,724 & 15.38\\
Gemma-7B   & 256,000 & 47,833 & 18.68  \\
Qwen2.5-7B & 152,064 & 18,973 & 12.48 \\
Amber-6.7B & 32,000 & 6,130   & 19.16 \\
\hline
\end{tabular}
\caption{
    The number of tokens in the vocabulary of each LLM (\textbf{\# Vocab}) and in our dataset (\textbf{\# Dataset}, \S \ref{sec:dataset}). \% shows the ratio of in-dataset tokens in the vocabulary.
}
\label{tbl:dataset}
\end{table}
Our evaluation focuses on spelling out \emph{single} tokens into their constituent characters. 
In other words, we exclude multi-token words to ensure a well-controlled experimental setting.
For example, in the case of ``\texttt{token/s}'', the model could easily reveal the final ``\texttt{s}'' without requiring character-level understanding.

We construct the dataset from each model’s vocabulary by selecting all single tokens that: 1) contain only lowercase alphabets (\texttt{a-z}), 2) begin with the special prefix indicating the head of words (e.g., ``\texttt{\_}'' in ``\texttt{\_token}''), and 3) are at least five characters. 
This yields a set comprising roughly 15 \% of each model’s full vocabulary (Table \ref{tbl:dataset}).  Note that some tokens (e.g., “somew,” “immedi”) are subword fragments rather than complete words.
Figure \ref{fgr:lengthFreq} and \ref{fgr:alphabetFreq} in the appendix show the distribution of frequency of token length and alphabet, respectively.

Given the target tokens in our dataset, LLMs are expected to output a sequence of compositional characters of each token.
In our experiment, we represent the spelling-out with a whitespace separation.
For example, LLMs with an input ``\texttt{token}'' need to generate ``\texttt{t o k e n}.''

\section{LLMs can Spell-out Tokens}
\label{sec:preliminary}

\begin{table}[t]
\centering
\small
\begin{tabular}{l:l}
\hline
\textbf{Few-shot Example}   & \texttt{hello : h e l l o,}    \\
                   & \texttt{world : w o r l d,}    \\
                   & \texttt{orange : o r a n g e,} \\
\textbf{Single Token Input} & \texttt{libert :}              \\
\textbf{Expected Output}    & \texttt{l i b e r t} \\
\hline
\end{tabular}
\caption{
    An input example for the three-shot setting.
}
\label{tbl:promptExample}
\end{table}
\begin{figure}[t]
\centering
\includegraphics[width=7.5cm]{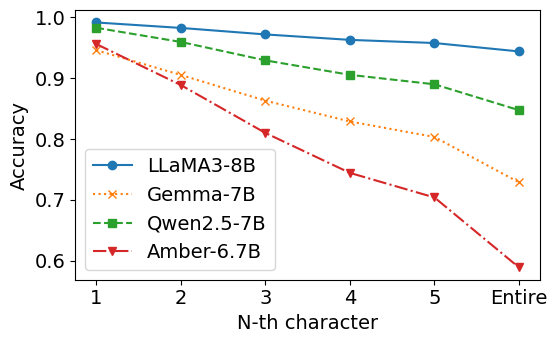}
\caption{
    Experimental results with the few-shot setting.
    $N=1...5$ means the accuracy of the $N$-th character prediction.
    ``Entire'' shows the accuracy of the prediction for all characters of tokens.
}
\label{fgr:fewshotResult}
\end{figure}

We first assess each LLM's ability to spell out single tokens given few-shot examples, following prior work on word-level spelling-out~\cite{edman-etal-2024-cute, xiong2025enhancing}. 
Table~\ref{tbl:promptExample} shows a prompt of a three-shot example for this experiment. 
We measure performance over the full evaluation dataset created in \S \ref{sec:dataset} for each LLM.
We count a prediction as correct only if the model’s output exactly matches the ground-truth character sequence with no missing or extra characters.

The data points named ``Entire'' in Figure~\ref{fgr:fewshotResult} report the overall token-level spelling-out accuracy of each LLM.
Although direct comparison across models is impossible because each uses its own vocabulary subset, we observe substantial differences: LLaMA3-8B achieves 94.41\% accuracy, whereas Amber-6.7B reaches only 58.86\%.
This variation shows the impact of model architecture and pretraining data on character-level capabilities.

In Figure~\ref{fgr:fewshotResult}, we also report accuracy by character index within each token. 
All models achieve over 94\% accuracy on the first character, but accuracy steadily declines for later positions.
This result indicates that correctly generating characters further along in the token becomes increasingly difficult.
Notably, however, every model maintains over 70\% accuracy through the fifth character, demonstrating LLMs' robust mid-token spelling capability.



\section{Embeddings do not Know Characters}
\label{sec:embedProbing}
While Section~\ref{sec:preliminary} demonstrated that LLMs can spell out single tokens, two key questions remain: where is the character-level knowledge stored in the model, and how is that knowledge utilized during spelling? 
A natural hypothesis is that token embeddings themselves encode this information, since the spelling of a token (i.e., its sequence of characters) is inherently context‐independent and embeddings are derived directly from token identities.
This section investigates whether the token embedding stores the knowledge of spelling tokens.

\subsection{MLP Probing with Token Embedding}
\begin{figure}[t]
\centering
\includegraphics[width=7.5cm]{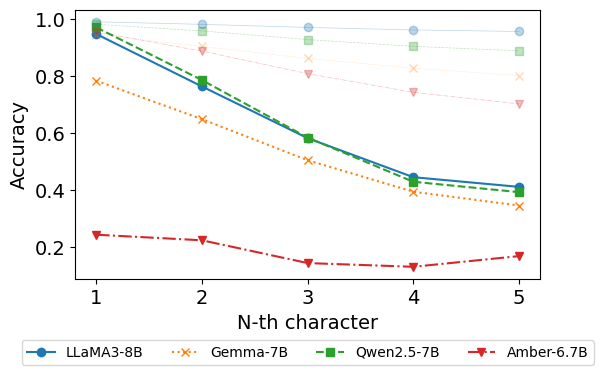}
\caption{
    The performance of probing classifiers for each $N$ (dark-colored lines).
    The light-colored lines show the few-shot performance copied from Figure \ref{fgr:fewshotResult}.
}
\label{fgr:embedProbingResult}
\end{figure}
To investigate where LLMs store character-level information, we train probing classifiers that predict the $N$-th character of a token $t$ from its embedding $\mathbf{v}_t \in \mathbb{R}^d$. 
We train a separate MLP classifier\footnote{We selected a non-linear probing because a linear classifier was incapable of extracting character-level information from embeddings.} for each character position $N$ ($1 \leq N \leq 5$), using an identical architecture across positions.

The MLP classifier maps the $d$-dimensional token embedding $\mathbf{v}_t$ to a 26-dimensional logits vector, corresponding to the 26 lowercase English characters. 
The probability that the $N$-th character of $t$ is $c$ is computed as:
\begin{equation}
p(t_N=c | t) = \mathrm{softmax}(f(\mathbf{v}_t; \theta_N))_c,
\label{eq:embedProbing}
\end{equation}
where $f: \mathbb{R}^d \rightarrow \mathbb{R}^{26}$ is the MLP classifier with three linear layers with the tanh activation, and $\theta_N$ is the set of trainable parameters for predicting the $N$-th character. 
The $\mathrm{softmax}(\cdot)_c$ operation extracts the probability assigned to character $c$.

We train each probing classifier on randomly sampled 90\% of the dataset (§\ref{sec:dataset}) and evaluate it on the remaining 10\%. 
This process is repeated ten times as $k$-fold cross-validation. 
If the classifier can accurately predict characters at a given position, we interpret this as evidence that the token embedding encodes character-level information at that position.
Training is performed using cross-entropy loss and the Adam~\cite{kingma2014adam} optimizer with a maximum of 300 epochs of training.
We used early stopping based on the training loss.

\subsection{Experimental Result}
Figure~\ref{fgr:embedProbingResult} shows the performance of the probing classifiers across character positions. 
For LLMs with larger vocabulary sizes (i.e., LLaMA3-8B, Gemma-7B, and Qwen2.5-7B), the classifier achieves over 80\% accuracy in predicting the first character. 
However, performance declines consistently as the character position increases. 
In parallel, the gap between probing accuracy (dark-colored lines) and few-shot accuracy (light-colored lines) widens at later positions.
These results suggest that LLMs rely on token embeddings to retrieve the first character, but access character-level information from upper layers when spelling out later characters.

Amber-6.7B, the model with the smallest vocabulary, shows a distinct trend. 
Its probing accuracy is significantly lower even for the first character.
This result indicates that its token embeddings carry little or no character-level information\footnote{Given that probing performance on intermediate layers of Amber is reasonable (§\ref{sec:layerProbe}), the lower embedding-level accuracy is unlikely to be caused by the smaller dataset size (Table~\ref{tbl:dataset}).}.

In summary, these results indicate that token embeddings in LLMs do not encode full character composition. 
While the first character is sometimes recoverable, character-level knowledge beyond that is primarily stored in the LLM's upper layers.

\begin{figure*}[t]
\centering
\includegraphics[width=\linewidth]{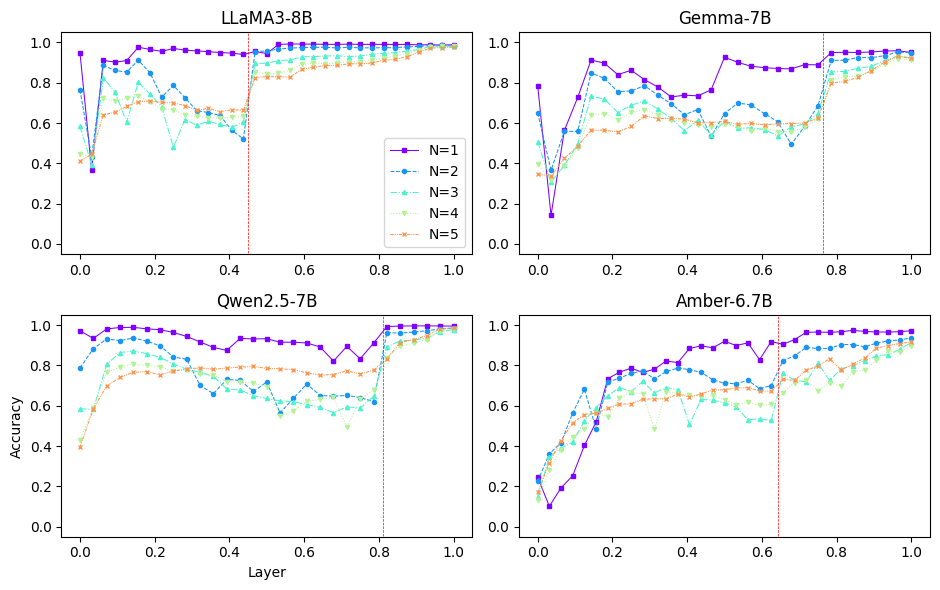}
\caption{
    Accuracy of $N$-th character prediction by probing classifiers at each Transformer layer.  
    The X-axis represents relative layer depth (0.0 = embedding layer, 1.0 = final layer).
    Red vertical lines indicate the breakthrough layers, which are calculated based on the average performance improvement between adjacent layers for $2 \leq N$.
}
\label{fgr:layerProbingResult}
\end{figure*}

\section{Which Layer Knows Spelling-out?}
\label{sec:layerProbe}
This section extends the probing analysis to internal Transformer layers to investigate where character-level knowledge emerges within the model.

\subsection{MLP Probing with Layer Output}
To analyze how character knowledge develops across the model, we apply the same probing classifier from §\ref{sec:embedProbing}, but instead of the token embedding $\mathbf{v}_t$, we use hidden states from individual Transformer layers.
Specifically, for predicting the $N$-th character, we extract the hidden state $\mathbf{h}^{l}_{N-1}$ from the $l$-th layer, corresponding to the token immediately before the character being predicted.
For example, to predict the third character ($N=3$) of the token ``\texttt{token},'' we extract the hidden state for the final token of the input sequence ``\texttt{[few-shot examples] token : t o}'', expecting the model to predict “\texttt{k}”.

All other aspects of the probing setup, including model architecture, training procedure, and evaluation, remain the same as described in \S \ref{sec:embedProbing}.

\subsection{Experimental Result}
Figure~\ref{fgr:layerProbingResult} presents the character prediction accuracy of probing classifiers across Transformer layers for all four LLMs. 
A consistent two-peak trend emerges: classifiers using early-layer hidden states can predict characters to some extent but not perfectly, whereas those using higher-layer representations can almost perfectly predict characters at all positions, following a performance drop in the intermediate layers.
The accuracy at the final layer closely matches that in Figure~\ref{fgr:fewshotResult}, supporting the validity of our evaluation setup.

Interestingly, LLaMA3-8B and Gemma-7B exhibit a notable dip in accuracy at the first Transformer layer (i.e., the second data point from the left). 
This suggests that character-level information is not immediately accessible after the embedding layer and may even be disrupted at this stage. 

For models other than Amber-6.7B, character prediction reaches higher accuracy in the early layers, but then temporarily declines before recovering in later layers. 
This suggests that LLMs do not simply pass character-level information through the network, but they engage in intermediate processing that reorganizes it as part of building the knowledge for solving the spelling-out task.

All models show a distinct ``breakthrough'' point, highlighted as red vertical lines, where the accuracy of the character prediction sharply increases in later layers. 
For example, LLaMA3-8B exhibits a jump in accuracy around layer depth 0.45. 
Amber-6.7B also shows an upward trend in the later layers, though the improvement is more modest compared to the other models.
This trend can be seen much more clearly when we use an explicit separator ``\texttt{/}'' for the spelling-out (Figure \ref{fgr:layerProbingResultSlash}).

These results suggest that LLMs begin to consolidate character-level spelling knowledge in the later stages of the network. 
In other words, the model first interprets the spelling-out task in its intermediate layers and then may act more like \textit{a character-level language model} in its final layers. 
This interpretation aligns with prior findings showing that LLM hidden states gradually shift toward representations resembling the next predicted token~\cite{voita-etal-2024-neurons,chang2025bigram}.

Finally, we observe that probing accuracy continues to rise even after the breakthrough point, particularly in the final two layers. 
This reinforces the view that character-level information is not statically stored in the embedding layer but is dynamically constructed and refined throughout the model, especially toward the end of the forward pass.


\begin{figure*}[t]
\centering
\includegraphics[width=\linewidth]{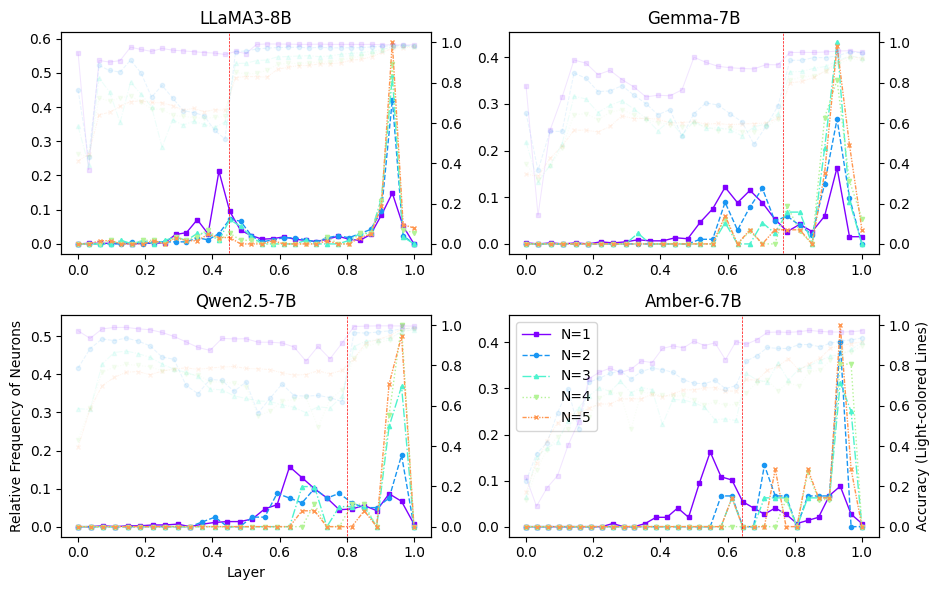}
\caption{
    Distribution of knowledge neurons for each layer (dark-colored lines).
    The accuracy of probing classifiers (light-colored lines) are copied from Figure \ref{fgr:layerProbingResult}.
}
\label{fgr:neuronDist}
\end{figure*}

\section{Knowledge Neuron for Spelling-out}
\label{sec:neuron}

The breakthrough point observed in \S \ref{sec:layerProbe} implies that large language models (LLMs) possess the ability to spell out tokens in the layers preceding this point. 
Recent work on interpretability has shown that individual neurons in Transformer models may encode factual knowledge or skills, as in knowledge neurons~\cite{dai2022knowledge} and skill neurons~\cite{wang2022finding}. 
Inspired by this line of research, we investigate where the knowledge for spelling out tokens resides within the neurons of each LLM.

\subsection{Knowledge Neurons}
In line with previous studies on neuron-level analysis of Transformer architectures~\cite{geva2021transformer}, we define knowledge neurons as the outputs of activation functions in the feed-forward network (FFN) sublayers of Transformer blocks. 
To quantify each neuron's contribution to a specific output character, we compute an attribution score.

For instance, given an input $x = $``\texttt{token : t o}'', we aim to measure how much a neuron $w$ contributes to predicting the next correct character $c= $``\texttt{k}''. 
We define the attribution score $\mathrm{Attr}(w|c, x)$ for the neuron $w$ using the following integrated gradient-based approximation:
\begin{equation}
\mathrm{Attr}(w|c, x) = \frac{\bar{w}}{m}\sum_{k=1}^{m}\frac{\partial P_{c,x}(\frac{k}{m}\bar{w})}{\partial w},
\end{equation}
where $\bar{w}$ is the activation value of neuron $w$ when the LLM processes the input $x$.
Here, $P_{c,x}(\bar{w})$ denotes the model's predicted probability of character $c$ given input $x$ when $w$ is  $\bar{w}$, and $m=20$ is the number of interpolation steps used in the Riemann sum approximation, as in~\newcite{dai2022knowledge}.

Using this attribution score, we identify knowledge neurons responsible for generating the $N$-th character in the spelling-out task.
In other words, we examine how neuron activations vary depending on the position of the predicted character.

To identify knowledge neurons for each character position $N$, we proceed as follows.
For each of the 1,000 sampled tokens from the dataset, we compute $\mathrm{Attr}(w|c,x)$ for all neurons and rank them.
Then, for each token, we select the top 1\% of neurons with the highest attribution scores at position $N$.
Finally, we define a neuron as a knowledge neuron for position $N$ if it appears in the top 1\% set of at least 75\% of the 1,000 tokens.

\subsection{Distribution of Neurons for $N$}

Figure~\ref{fgr:neuronDist} shows the distribution of knowledge neurons across layers for predicting the $N$-th character.
The distributions commonly exhibit two peaks: one at intermediate layers and another near the final layers.
However, the peak for the first character tends to occur in intermediate layers, while the peaks for later characters (e.g., $N=2$ and beyond) appear more prominently near the final layers.
This trend is consistent across all four models, suggesting a general phenomenon shared by various LLMs.

We also observe that the location of the first peak is loosely aligned with the breakthrough point identified earlier.
For example, in LLaMA3-8B, where the breakthrough occurs around mid-depth (layer $\sim 0.5$), the first peak of knowledge neurons also appears around the same depth.
Conversely, in Gemma-7B, where the breakthrough appears later, the first peak of neurons shifts to a correspondingly later layer.
These findings suggest that knowledge neurons in the mid-depth layers may play a foundational role in the spelling-out process.

\subsection{Intersection of Neurons}
\label{sec:intersection}
\begin{figure}[t]
\centering
\includegraphics[width=6.0cm]{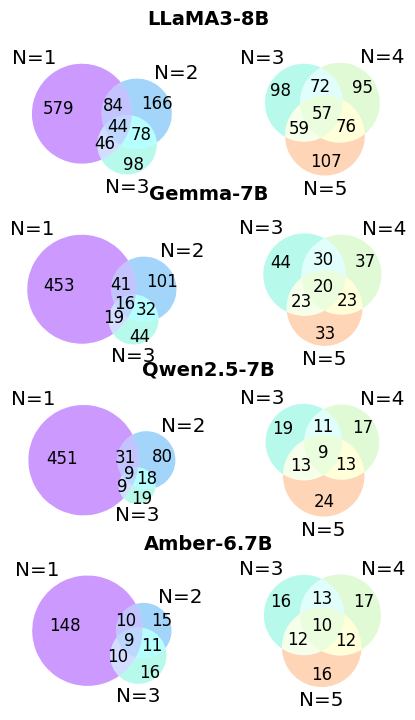}
\caption{
    Venn diagrams showing the number of overlapping knowledge neurons across different character positions $N$ for four LLMs.
}
\label{fgr:venn}
\end{figure}

Figure~\ref{fgr:venn} illustrates the overlap of knowledge neurons across different character positions $N$.
Across all models, the largest number of neurons is uniquely identified for the first character prediction, in contrast to those for the second and third characters, which share more neurons.

Moreover, the total number of knowledge neurons tends to decrease from $N=1$ to $N=3$.
This suggests that LLMs use a broader and more redundant set of neurons when initiating the spelling-out process, but rely on fewer and more specialized neurons as the character position progresses.

For later positions ($N=3$, $4$, $5$), both the number of knowledge neurons and the extent of their overlap remain relatively small and stable across all models.
This suggests a shift in internal strategy: while LLMs rely on a broad and shared set of neurons to initiate spelling-out at $N=1$, they gradually move toward using more position-specific and case-dependent neurons for later characters.
In other words, the model appears to generalize the beginning of spelling with common mechanisms, but adapts to the unique structure of each token as the character position advances.

These consistent trends across the four LLMs support a general conclusion:
LLMs tend to rely on a broad and shared set of neurons for predicting the first character, and to some extent the second, using a general mechanism to initiate the spelling-out process.
In contrast, for characters beyond the third position, they employ more case-specific neurons, indicating a shift toward more specialized processing tailored to individual token structures.


\subsection{Neuron Ablation}
To gain deeper insight into the roles of individual neurons, we ablated the top 100 most influential neurons for each character position $N$ and measured the resulting performance degradation relative to the original accuracy shown in Figure \ref{fgr:fewshotResult}.
We used LLaMA3-8B for this experiment, as it provides a sufficient number of neurons for meaningful ablation analysis (Figure \ref{fgr:venn}).

\begin{figure}[t]
\centering
\includegraphics[width=7.5cm]{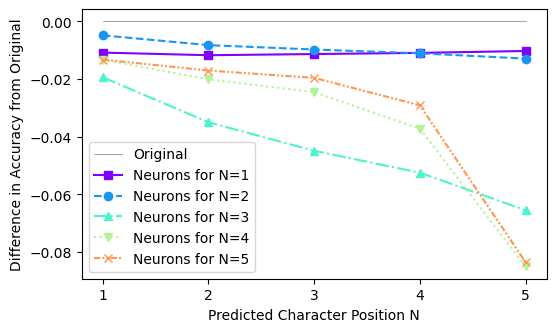}
\caption{
    Difference in accuracy with ablating 100 neurons for each character position $N$ (LLaMA3-8B).
}
\label{fgr:intervention}
\end{figure}

The experimental results in Figure \ref{fgr:intervention} show that ablating neurons for $N = {1, 2}$ leads to only modest accuracy loss, whereas ablating neurons for $N \geq 3$ causes a more substantial drop in performance.
This suggests that later-position neurons play a more critical role in the spelling-out task, compensating for the functions of earlier ones by leveraging contextual information.
These results support our hypothesis in Section~\ref{sec:intersection}: early-position neurons primarily extract features from initial character embeddings, whereas later-position neurons are responsible for context-based character prediction.

\section{Attention Weight for Spelling-out}
\label{sec:attention}
\begin{figure*}[t]
\centering
\includegraphics[width=\linewidth]{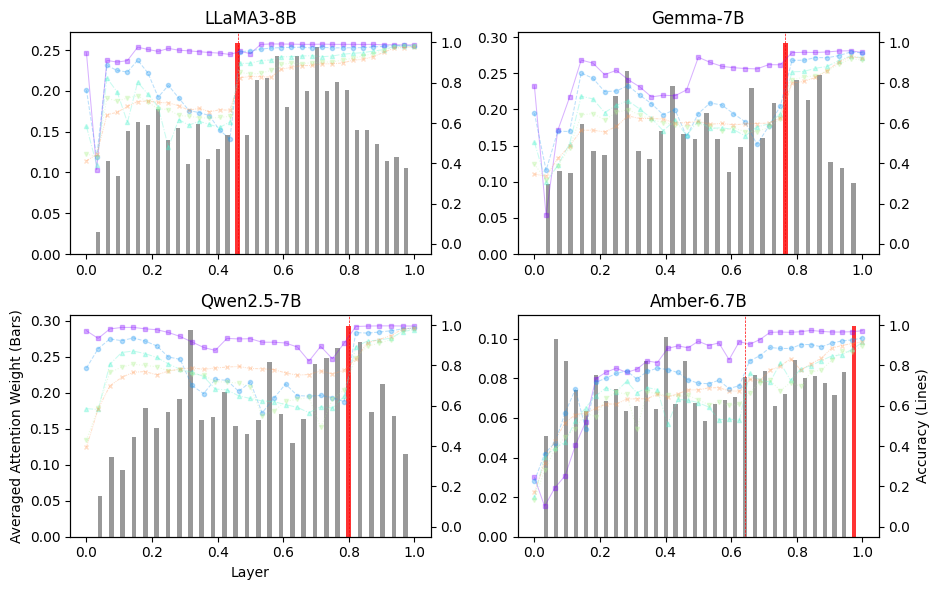}
\caption{
    Averaged attention weights to the target token across layers (bar graph).
    The red bar indicates the layer with the highest attention weight.
    The line graph shows the performance of predicting the $N$-th character using the probing classifier, replicated from Figure \ref{fgr:layerProbingResult}.
}
\label{fgr:attention}
\end{figure*}

Given the nature of the spelling-out task, LLMs must attend to the target token while generating its characters.
The ability to focus attention on the correct token is crucial for accurate spelling-out, and we hypothesize that this ability corresponds to the breakthrough identified in \S \ref{sec:layerProbe}.
This section further investigates the relationship between spelling-out capability and attention weights to the target token.

Given a sequence consisting of the target token and its spelling-out, we compute the average attention weight from all related elements to the target token.
For example, when the input is \texttt{[few-shot examples] token : t o k e n}, where the target token is ``\texttt{token}'' following a few-shot example, we calculate the average attention weight to ``\texttt{token}'' from each of the elements: \texttt{token}, \texttt{:}, \texttt{t}, \texttt{o}, \texttt{k}, \texttt{e}, and \texttt{n}.
We average these attention weights over 1,000 samples and across all attention heads in each layer\footnote{These tokens are selected from the dataset as those that each LLM can correctly spell out in the few-shot setting (\S \ref{sec:preliminary}).}.
To isolate attention to the target token, we remove attention weights to ``\texttt{<BOS>}'' and renormalize the distributions prior to averaging, inspired by \citet{kobayashi-etal-2020-attention}.

Figure \ref{fgr:attention} presents the average attention scores across the 1,000 examples.
Interestingly, the layer with the highest attention to the target token (red bars) coincides with the breakthrough point (red lines) in three out of four models.
This finding suggests that the performance improvement observed in intermediate layers is due to the model's increasing ability to correctly attend to the target token.

The distinct result observed in Amber-6.7B suggests that performance breakthroughs do not necessarily align with the behavior of attention weights. 
Given the overall lower attention weights assigned to the target token compared to other models, we consider that this model needs to attend to broader contextual information rather than focusing on the single target token, possibly due to a lack of character-level knowledge in its embeddings.



\section{Conclusion}
This paper investigated the paradox that, although large language models (LLMs) struggle to recognize individual characters within words, they can accurately spell out words character by character.

Our probing analyses revealed that the token embedding layer does not encode complete character-level information, especially beyond the first character. 
Instead, character-level features are dynamically reconstructed in the intermediate and later layers, where we observe a distinct ``breakthrough'' point. Around this stage, models begin to reliably access and utilize character-level knowledge.

We further examined the spelling-out behavior through the lens of knowledge neurons and attention patterns, both of which align with the breakthrough layers. 
Crucially, this behavior suggests that spelling out is a learned task (i.e., dependent on identifying the target token and retrieving character-level information), rather than a simple extraction of character-level information stored in the embedding layers.

As a result, our findings imply that the apparent success of LLMs in spelling out does not extend to more complex or unfamiliar tasks such as reverse spelling, letter insertion, or character-level reasoning in novel contexts. 
Our findings indicate that current LLMs are not inherently character-aware; rather, they rely on task-specific heuristics acquired during training or prompting. 
This suggests that models must explicitly learn how to apply character knowledge to more complex manipulations.




\section*{Limitations}
While we believe that our experiments were conducted under well-controlled conditions and provide sufficient support for our hypotheses, we acknowledge the following limitations:

\begin{itemize}
    \item To ensure well-controlled experimental conditions, we restricted the target vocabulary to single-token words composed of lowercased alphabetic characters. Different trends may emerge when using other languages. However, considering prior work showing that LLMs can predict the initial radicals of Chinese characters~\cite{wu-etal-2025-impact}, we expect similar tendencies to hold across languages.
    \item This study investigates LLM behavior using probing classifiers, knowledge neurons, and attention heads. It is important to note, however, that these methods do not fully capture or explain the underlying mechanisms of model knowledge and capabilities~\cite{jain-wallace-2019-attention,belinkov-2022-probing,kumar2022probing,niudoes}.
    \item Our experiments focus on spelling-out behavior using whitespace as a separator. Although similar trends were observed when using an alternative separator (``\texttt{/}'', see Figure~\ref{fgr:layerProbingResultSlash}), the results may vary depending on input formatting.
    \item Our findings do not necessarily generalize to all LLMs. In fact, Amber-6.7B exhibited divergent behavior across several experiments, deviating from the patterns observed in other models. Nevertheless, we argue that presenting such counterexamples is essential for a deeper understanding of model behavior. In this work, we attribute Amber's deviation to its apparent lack of character-level information in the embedding layer, a distinctive property not shared by the other models. Given that the remaining three models consistently followed the same trends, we consider Amber-6.7B to be a special case rather than a representative counterpoint.

\end{itemize}



\bibliography{custom,anthology}

\begin{thebibliography}{31}
\providecommand{\natexlab}[1]{#1}

\bibitem[{Belinkov(2022)}]{belinkov-2022-probing}
Yonatan Belinkov. 2022.
\newblock \href {https://doi.org/10.1162/coli_a_00422} {Probing classifiers: Promises, shortcomings, and advances}.
\newblock \emph{Computational Linguistics}, 48(1):207--219.

\bibitem[{Chai et~al.(2024)Chai, Fang, Peng, and Li}]{chai-etal-2024-tokenization}
Yekun Chai, Yewei Fang, Qiwei Peng, and Xuhong Li. 2024.
\newblock \href {https://doi.org/10.18653/v1/2024.findings-emnlp.86} {Tokenization falling short: On subword robustness in large language models}.
\newblock In \emph{Findings of the Association for Computational Linguistics: EMNLP 2024}, pages 1582--1599, Miami, Florida, USA. Association for Computational Linguistics.

\bibitem[{Chang and Bergen(2025)}]{chang2025bigram}
Tyler~A Chang and Benjamin~K Bergen. 2025.
\newblock Bigram subnetworks: Mapping to next tokens in transformer language models.
\newblock \emph{arXiv preprint arXiv:2504.15471}.

\bibitem[{Dai et~al.(2022)Dai, Dong, Hao, Sui, Chang, and Wei}]{dai2022knowledge}
Damai Dai, Li~Dong, Yaru Hao, Zhifang Sui, Baobao Chang, and Furu Wei. 2022.
\newblock Knowledge neurons in pretrained transformers.
\newblock In \emph{Proceedings of the 60th Annual Meeting of the Association for Computational Linguistics (Volume 1: Long Papers)}, pages 8493--8502.

\bibitem[{Dubey et~al.(2024)Dubey, Jauhri, Pandey, Kadian, Al-Dahle, Letman, Mathur, Schelten, Yang, Fan et~al.}]{dubey2024llama}
Abhimanyu Dubey, Abhinav Jauhri, Abhinav Pandey, Abhishek Kadian, Ahmad Al-Dahle, Aiesha Letman, Akhil Mathur, Alan Schelten, Amy Yang, Angela Fan, et~al. 2024.
\newblock The llama 3 herd of models.
\newblock \emph{arXiv preprint arXiv:2407.21783}.

\bibitem[{Edman et~al.(2024)Edman, Schmid, and Fraser}]{edman-etal-2024-cute}
Lukas Edman, Helmut Schmid, and Alexander Fraser. 2024.
\newblock \href {https://doi.org/10.18653/v1/2024.emnlp-main.177} {{CUTE}: Measuring {LLM}s' understanding of their tokens}.
\newblock In \emph{Proceedings of the 2024 Conference on Empirical Methods in Natural Language Processing}, pages 3017--3026, Miami, Florida, USA. Association for Computational Linguistics.

\bibitem[{Fu et~al.(2024)Fu, Ferrando, Conde, Arriaga, and Reviriego}]{fu2024large}
Tairan Fu, Raquel Ferrando, Javier Conde, Carlos Arriaga, and Pedro Reviriego. 2024.
\newblock Why do large language models (llms) struggle to count letters?
\newblock \emph{arXiv preprint arXiv:2412.18626}.

\bibitem[{Geva et~al.(2021)Geva, Schuster, Berant, and Levy}]{geva2021transformer}
Mor Geva, Roei Schuster, Jonathan Berant, and Omer Levy. 2021.
\newblock Transformer feed-forward layers are key-value memories.
\newblock In \emph{Proceedings of the 2021 Conference on Empirical Methods in Natural Language Processing}, pages 5484--5495.

\bibitem[{Hiraoka and Okazaki(2024)}]{hiraoka2024knowledge}
Tatsuya Hiraoka and Naoaki Okazaki. 2024.
\newblock Knowledge of pretrained language models on surface information of tokens.
\newblock \emph{arXiv preprint arXiv:2402.09808}.

\bibitem[{Jain and Wallace(2019)}]{jain-wallace-2019-attention}
Sarthak Jain and Byron~C. Wallace. 2019.
\newblock \href {https://doi.org/10.18653/v1/N19-1357} {{A}ttention is not {E}xplanation}.
\newblock In \emph{Proceedings of the 2019 Conference of the North {A}merican Chapter of the Association for Computational Linguistics: Human Language Technologies, Volume 1 (Long and Short Papers)}, pages 3543--3556, Minneapolis, Minnesota. Association for Computational Linguistics.

\bibitem[{Kamoda et~al.(2025)Kamoda, Heinzerling, Inaba, Kudo, Sakaguchi, and Inui}]{kamoda-etal-2025-weight}
Go~Kamoda, Benjamin Heinzerling, Tatsuro Inaba, Keito Kudo, Keisuke Sakaguchi, and Kentaro Inui. 2025.
\newblock \href {https://aclanthology.org/2025.findings-naacl.355/} {Weight-based analysis of detokenization in language models: Understanding the first stage of inference without inference}.
\newblock In \emph{Findings of the Association for Computational Linguistics: NAACL 2025}, pages 6324--6343, Albuquerque, New Mexico. Association for Computational Linguistics.

\bibitem[{Kaplan et~al.(2025)Kaplan, Oren, Reif, and Schwartz}]{kaplan2025from}
Guy Kaplan, Matanel Oren, Yuval Reif, and Roy Schwartz. 2025.
\newblock \href {https://openreview.net/forum?id=328vch6tRs} {From tokens to words: On the inner lexicon of {LLM}s}.
\newblock In \emph{The Thirteenth International Conference on Learning Representations}.

\bibitem[{Kingma(2014)}]{kingma2014adam}
Diederik~P Kingma. 2014.
\newblock Adam: A method for stochastic optimization.
\newblock \emph{arXiv preprint arXiv:1412.6980}.

\bibitem[{Kobayashi et~al.(2020)Kobayashi, Kuribayashi, Yokoi, and Inui}]{kobayashi-etal-2020-attention}
Goro Kobayashi, Tatsuki Kuribayashi, Sho Yokoi, and Kentaro Inui. 2020.
\newblock \href {https://doi.org/10.18653/v1/2020.emnlp-main.574} {Attention is not only a weight: Analyzing transformers with vector norms}.
\newblock In \emph{Proceedings of the 2020 Conference on Empirical Methods in Natural Language Processing (EMNLP)}, pages 7057--7075, Online. Association for Computational Linguistics.

\bibitem[{Kumar et~al.(2022)Kumar, Tan, and Sharma}]{kumar2022probing}
Abhinav Kumar, Chenhao Tan, and Amit Sharma. 2022.
\newblock Probing classifiers are unreliable for concept removal and detection.
\newblock \emph{Advances in Neural Information Processing Systems}, 35:17994--18008.

\bibitem[{Liu et~al.()Liu, Qiao, Neiswanger, Wang, Tan, Tao, Li, Wang, Sun, Pangarkar et~al.}]{liullm360}
Zhengzhong Liu, Aurick Qiao, Willie Neiswanger, Hongyi Wang, Bowen Tan, Tianhua Tao, Junbo Li, Yuqi Wang, Suqi Sun, Omkar Pangarkar, et~al.
\newblock Llm360: Towards fully transparent open-source llms.
\newblock In \emph{First Conference on Language Modeling}.

\bibitem[{Marco and Fraser(2024)}]{marco-fraser-2024-subword}
Marion~Di Marco and Alexander Fraser. 2024.
\newblock \href {https://doi.org/10.18653/v1/2024.emnlp-main.672} {Subword segmentation in {LLM}s: Looking at inflection and consistency}.
\newblock In \emph{Proceedings of the 2024 Conference on Empirical Methods in Natural Language Processing}, pages 12050--12060, Miami, Florida, USA. Association for Computational Linguistics.

\bibitem[{Niu et~al.()Niu, Liu, Zhu, and Penn}]{niudoes}
Jingcheng Niu, Andrew Liu, Zining Zhu, and Gerald Penn.
\newblock What does the knowledge neuron thesis have to do with knowledge?
\newblock In \emph{The Twelfth International Conference on Learning Representations}.

\bibitem[{Shin and Kaneko(2024)}]{shin2024large}
Andrew Shin and Kunitake Kaneko. 2024.
\newblock Large language models lack understanding of character composition of words.
\newblock In \emph{ICML 2024 Workshop on LLMs and Cognition}.

\bibitem[{Team et~al.(2024)Team, Mesnard, Hardin, Dadashi, Bhupatiraju, Pathak, Sifre, Rivière, Kale, Love et~al.}]{team2024gemma}
Gemma Team, Thomas Mesnard, Cassidy Hardin, Robert Dadashi, Surya Bhupatiraju, Shreya Pathak, Laurent Sifre, Morgane Rivière, Mihir~Sanjay Kale, Juliette Love, et~al. 2024.
\newblock \href {https://arxiv.org/abs/2403.08295} {Gemma: Open models based on gemini research and technology}.

\bibitem[{Tsuji et~al.(2025)Tsuji, Hiraoka, Cheng, Aramaki, and Iwakura}]{tsuji2025investigating}
Kohei Tsuji, Tatsuya Hiraoka, Yuchang Cheng, Eiji Aramaki, and Tomoya Iwakura. 2025.
\newblock Investigating neurons and heads in transformer-based llms for typographical errors.
\newblock \emph{arXiv preprint arXiv:2502.19669}.

\bibitem[{Vaswani et~al.(2017)Vaswani, Shazeer, Parmar, Uszkoreit, Jones, Gomez, Kaiser, and Polosukhin}]{vaswani2017attention}
Ashish Vaswani, Noam Shazeer, Niki Parmar, Jakob Uszkoreit, Llion Jones, Aidan~N Gomez, \L~ukasz Kaiser, and Illia Polosukhin. 2017.
\newblock \href {https://proceedings.neurips.cc/paper_files/paper/2017/file/3f5ee243547dee91fbd053c1c4a845aa-Paper.pdf} {Attention is all you need}.
\newblock In \emph{Advances in Neural Information Processing Systems}, volume~30. Curran Associates, Inc.

\bibitem[{Voita et~al.(2024)Voita, Ferrando, and Nalmpantis}]{voita-etal-2024-neurons}
Elena Voita, Javier Ferrando, and Christoforos Nalmpantis. 2024.
\newblock \href {https://doi.org/10.18653/v1/2024.findings-acl.75} {Neurons in large language models: Dead, n-gram, positional}.
\newblock In \emph{Findings of the Association for Computational Linguistics: ACL 2024}, pages 1288--1301, Bangkok, Thailand. Association for Computational Linguistics.

\bibitem[{Wang et~al.(2025)Wang, Gu, Wei, Gao, Song, and Chen}]{wang2025word}
Chenxi Wang, Tianle Gu, Zhongyu Wei, Lang Gao, Zirui Song, and Xiuying Chen. 2025.
\newblock Word form matters: Llms' semantic reconstruction under typoglycemia.
\newblock \emph{arXiv preprint arXiv:2503.01714}.

\bibitem[{Wang et~al.(2022)Wang, Wen, Zhang, Hou, Liu, and Li}]{wang2022finding}
Xiaozhi Wang, Kaiyue Wen, Zhengyan Zhang, Lei Hou, Zhiyuan Liu, and Juanzi Li. 2022.
\newblock Finding skill neurons in pre-trained transformer-based language models.
\newblock In \emph{Proceedings of the 2022 Conference on Empirical Methods in Natural Language Processing}, pages 11132--11152.

\bibitem[{Wang et~al.(2024)Wang, Fu, Wang, and Gong}]{wang2024stringllm}
Xilong Wang, Hao Fu, Jindong Wang, and Neil~Zhenqiang Gong. 2024.
\newblock Stringllm: Understanding the string processing capability of large language models.
\newblock \emph{arXiv preprint arXiv:2410.01208}.

\bibitem[{Wolf et~al.(2020)Wolf, Debut, Sanh, Chaumond, Delangue, Moi, Cistac, Rault, Louf, Funtowicz, Davison, Shleifer, von Platen, Ma, Jernite, Plu, Xu, Le~Scao, Gugger, Drame, Lhoest, and Rush}]{wolf2020transformers}
Thomas Wolf, Lysandre Debut, Victor Sanh, Julien Chaumond, Clement Delangue, Anthony Moi, Pierric Cistac, Tim Rault, Remi Louf, Morgan Funtowicz, Joe Davison, Sam Shleifer, Patrick von Platen, Clara Ma, Yacine Jernite, Julien Plu, Canwen Xu, Teven Le~Scao, Sylvain Gugger, Mariama Drame, Quentin Lhoest, and Alexander Rush. 2020.
\newblock \href {https://doi.org/10.18653/v1/2020.emnlp-demos.6} {Transformers: State-of-the-art natural language processing}.
\newblock In \emph{Proceedings of the 2020 Conference on Empirical Methods in Natural Language Processing: System Demonstrations}, pages 38--45, Online. Association for Computational Linguistics.

\bibitem[{Wu et~al.(2025)Wu, Stratos, and Xu}]{wu-etal-2025-impact}
Xiaofeng Wu, Karl Stratos, and Wei Xu. 2025.
\newblock \href {https://aclanthology.org/2025.naacl-long.16/} {The impact of visual information in {C}hinese characters: Evaluating large models' ability to recognize and utilize radicals}.
\newblock In \emph{Proceedings of the 2025 Conference of the Nations of the Americas Chapter of the Association for Computational Linguistics: Human Language Technologies (Volume 1: Long Papers)}, pages 331--350, Albuquerque, New Mexico. Association for Computational Linguistics.

\bibitem[{Xiong et~al.(2025)Xiong, Cai, Hooi, Peng, Li, and Wang}]{xiong2025enhancing}
Zhen Xiong, Yujun Cai, Bryan Hooi, Nanyun Peng, Zhecheng Li, and Yiwei Wang. 2025.
\newblock Enhancing llm character-level manipulation via divide and conquer.
\newblock \emph{arXiv preprint arXiv:2502.08180}.

\bibitem[{Xu et~al.(2024)Xu, Zhao, Zhang, Liu, Shen, Liu, Kuang, He, and Liu}]{xu2024enhancing}
Zhu Xu, Zhiqiang Zhao, Zihan Zhang, Yuchi Liu, Quanwei Shen, Fei Liu, Yu~Kuang, Jian He, and Conglin Liu. 2024.
\newblock Enhancing character-level understanding in llms through token internal structure learning.
\newblock \emph{arXiv preprint arXiv:2411.17679}.

\bibitem[{Yang et~al.(2024)Yang, Yang, Zhang, Hui, Zheng, Yu, Li, Liu, Huang, Wei et~al.}]{yang2024qwen2}
An~Yang, Baosong Yang, Beichen Zhang, Binyuan Hui, Bo~Zheng, Bowen Yu, Chengyuan Li, Dayiheng Liu, Fei Huang, Haoran Wei, et~al. 2024.
\newblock Qwen2. 5 technical report.
\newblock \emph{arXiv preprint arXiv:2412.15115}.

\end{thebibliography}

\newpage
\appendix

\section{Experimental Environment}
All experiments in this paper were conducted using the HuggingFace Transformers library~\cite{wolf2020transformers}.
Unless otherwise specified, we used default hyperparameter settings.
Most experiments were performed on NVIDIA A40 GPUs, with the exception of those involving Gemma-7B, which were conducted on NVIDIA H100 GPUs.

The most computationally intensive part of our work was the identification of neurons described in \S\ref{sec:neuron}, which took over 24 hours to process 1,000 samples.
All other experiments were completed within 24 hours using a single GPU.

\section{Additional Dataset Statistics}
\label{sec:appendix}

\begin{figure}[t]
\centering
\includegraphics[width=7.5cm]{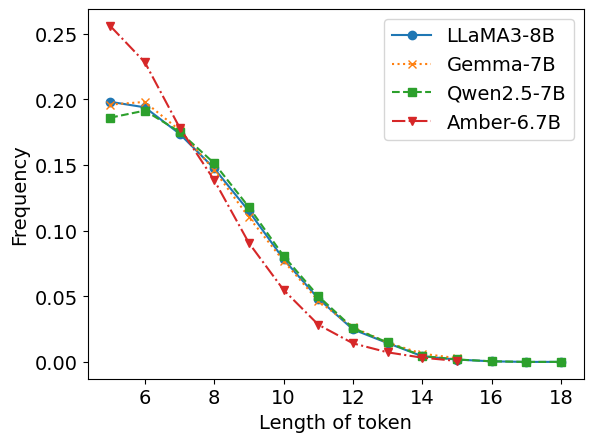}
\caption{
    Token length frequency distribution in the dataset.
}
\label{fgr:lengthFreq}
\end{figure}

\begin{figure}[t]
\centering
\includegraphics[width=7.5cm]{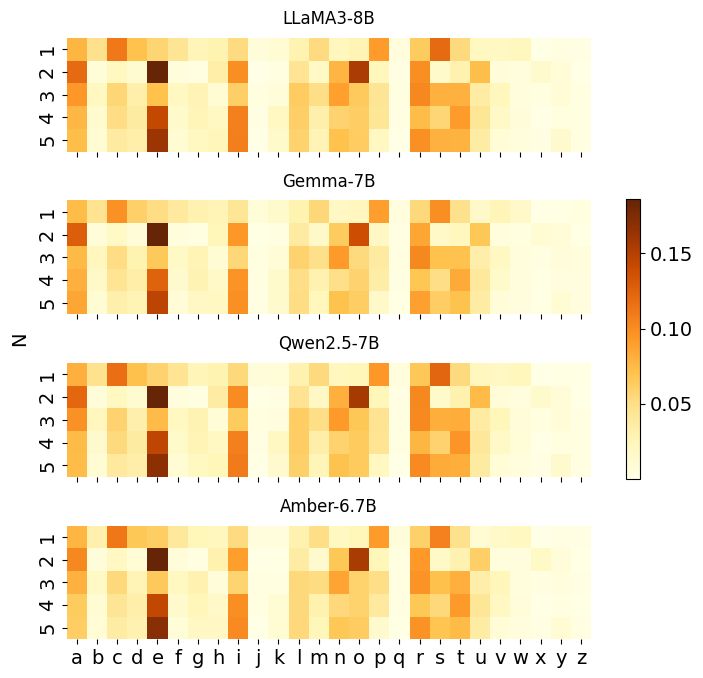}
\caption{
    Alphabet frequency distribution at each character position in the dataset.
}
\label{fgr:alphabetFreq}
\end{figure}

Figure~\ref{fgr:lengthFreq} shows the distribution of token lengths in our dataset.
As illustrated, the three LLMs with larger vocabularies exhibit similar distributions.
Figure~\ref{fgr:alphabetFreq} presents the distribution of characters (alphabets) at each position in the tokens, showing highly consistent patterns across models.

\section{Few-shot Results by Token Length}
\begin{figure}[t]
\centering
\includegraphics[width=7.5cm]{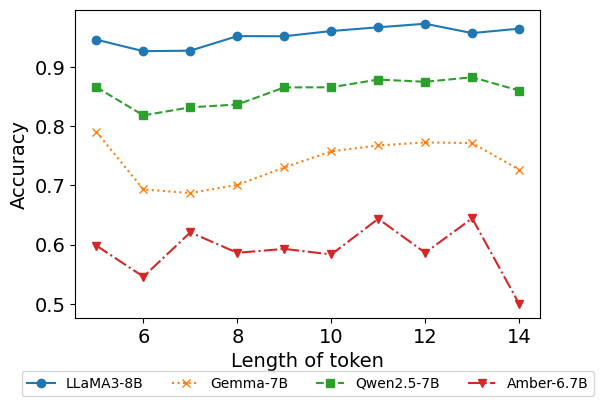}
\caption{
    Few-shot spelling-out accuracy by input token length.
}
\label{fgr:perLength}
\end{figure}

Figure~\ref{fgr:perLength} plots the few-shot spelling-out accuracy from \S\ref{sec:preliminary} as a function of token length (5–14 characters).
Contrary to the intuition that longer tokens would be harder to spell out correctly, the relatively flat trend suggests that token length has a limited impact on exact token-level accuracy.

\section{Layer Probing with ``/'' Separation}
\begin{figure*}[t]
\centering
\includegraphics[width=\linewidth]{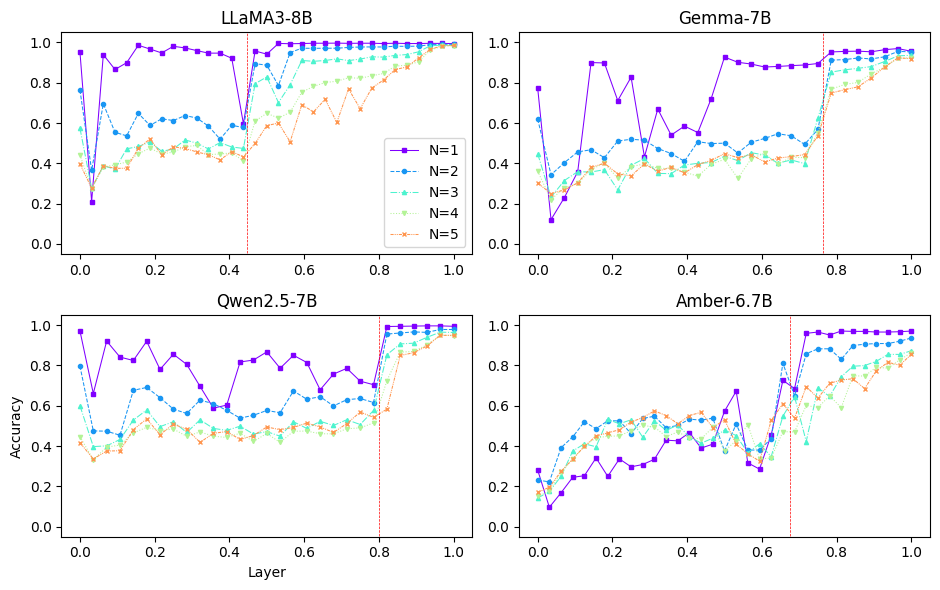}
\caption{
    Accuracy of $N$-th character prediction using the ``\texttt{/}'' separator, measured via probing classifiers across Transformer layers.
}
\label{fgr:layerProbingResultSlash}
\end{figure*}

Figure~\ref{fgr:layerProbingResultSlash} presents the results of probing classifiers when the separator used for spelling out is changed from whitespace to a forward slash (``\texttt{/}''). 
The experimental setup remains the same as in \S\ref{sec:layerProbe}, except for this separator change. 
This modified separator was used consistently across few-shot examples, target inputs, and expected outputs (see Figure~\ref{tbl:promptExample}). 
For instance, the input ``\texttt{hello :/h/e/l/l/o/}'' was used in one of the examples.

To ensure consistency of representation across character positions, we added a slash before and after the spelling so that the model does not need to use tokens with the special prefix indicating the word head.
In contrast, the main experiments used inputs such as ``\texttt{\_h\_e\_l\_l\_o}'', where the underscore (``\texttt{\_}'') denotes a prefix.

As shown in Figure~\ref{fgr:layerProbingResultSlash}, we observe a similar trend to that in Figure~\ref{fgr:layerProbingResult}, including the occurrence of a ``breakthrough'' at approximately the same layer.
This consistency suggests that our findings regarding breakthrough behavior may generalize across different input formats for spelling tasks.

\section{Neurons for Alphabet}
\begin{figure*}[t]
\centering
\includegraphics[width=\linewidth]{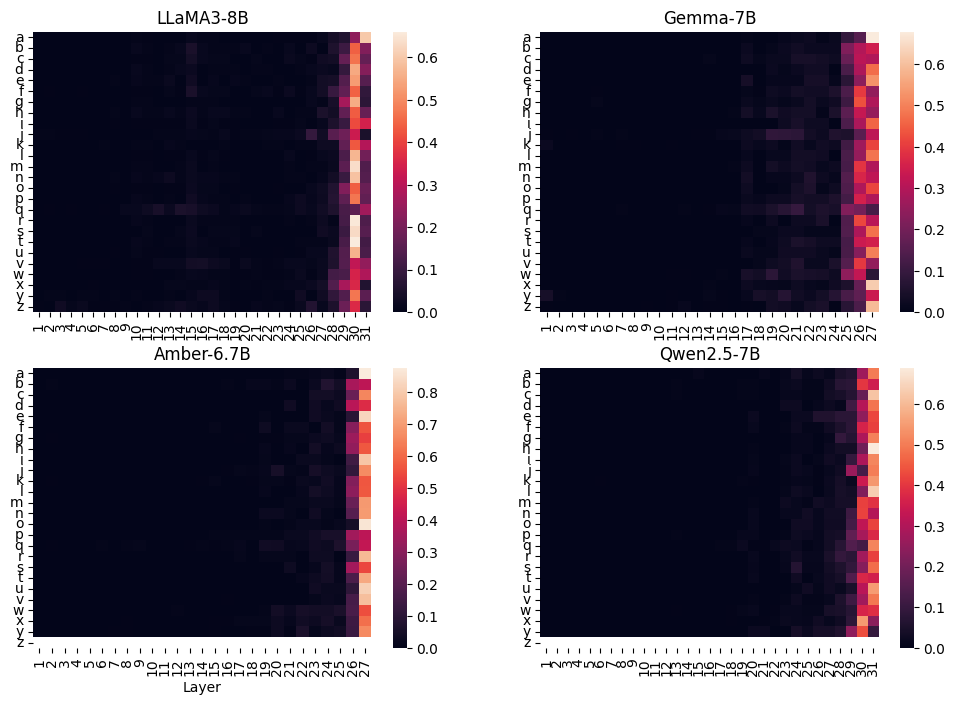}
\caption{
    Distribution of knowledge neurons responsible for outputting each alphabet character.
}
\label{fgr:neuron_alphabet}
\end{figure*}

Figure~\ref{fgr:neuron_alphabet} shows the distribution of knowledge neurons associated with the output of each alphabet character.
We used the same identification method as described in \S\ref{sec:neuron}, but focused on individual characters rather than positional indices ($N$).
The results indicate that neurons responsible for alphabet outputs are primarily located in the near-final layers of the models.

\end{document}